\documentclass[twoside]{article}
\usepackage{nips11submit_e,times}

\usepackage{graphicx}

\usepackage{latexsym} 
\usepackage{amsmath}

\usepackage{hyperref}

\nipsfinalcopy 

\newcommand{\ewaweight}{\phi}
\newcommand{\fsweight}{w}

\newcommand{\ensemblesize}{q}

\newtheorem{theorem}{Theorem}
\newtheorem{corollary}{Corollary}

\title{Adapting to Non-stationarity with Growing Expert Ensembles}
\author{
Cosma Rohilla Shalizi\\
Carnegie Mellon University\\
Pittsburgh, PA\\
\texttt{cshalizi@cmu.edu}\\
\And
Abigail Z. Jacobs\\
Northwestern University\\ 
Evanston, IL\\
\texttt{azjacobs@northwestern.edu}\\ 
\And
Kristina Lisa Klinkner\\
Amazon.com\\
Seattle, WA\\
\texttt{klinkner@amazon.com}\\
\And
Aaron Clauset\\
University of Colorado\\
Boulder, CO\\
\texttt{aaron.clauset@colorado.edu}\\
}

\begin{document}

\maketitle

\begin{abstract}
  When dealing with time series with complex non-stationarities, low
  retrospective regret on individual realizations is a more appropriate goal
  than low prospective risk in expectation.  Online learning algorithms provide
  powerful guarantees of this form, and have often been proposed for use with
  non-stationary processes because of their ability to switch between different
  forecasters or ``experts''.  However, existing methods assume that the set of
  experts whose forecasts are to be combined are all given at the start, which
  is not plausible when dealing with a genuinely historical or evolutionary
  system.  We show how to modify the ``fixed shares'' algorithm for tracking
  the best expert to cope with a steadily growing set of experts, obtained by
  fitting new models to new data as it becomes available, and obtain regret
  bounds for the growing ensemble.
\end{abstract}

\section{Introduction}

Non-stationarity is ubiquitous in the study of real time series; macroeconomic
statistics, climate records and gene expression levels are all prominent
examples, as are important engineering problems of signal processing and
anomaly detection.  Sometimes the nonstationarity is harmless, as when the data
come from a homogeneous Markov process, or more generally from a conditionally
stationary \cite{Caires-Ferreira-prediction-of-cond-stat-seq} source, since
then the best prediction for each historical context is invariant, though
various contexts become more or less common.  More generally, however,
non-stationary processes have trends, so the predictive implications of any
given historical context changes over time.

Time series textbooks (such as \cite{Shumway-Stoffer}) advise turning
non-stationary processes into stationary ones, by, e.g., subtracting off trends
and then analyzing the residuals as a stationary process.  If there are
multiple independent replicas of the process, all with the same trend, the
latter could be estimated non-parametrically.  If there is only one
realization, systematically estimating the trend needs a well-specified
parametric model embracing both trend and fluctuations.  Time series from
complex systems, however, typically lack parametric models of trends deserving
much credence.  In macroeconomics, for instance, the state-of-the-art models
are all for stationary fluctuations, and trends are identified by {\em ad hoc}
procedures, most often spline smoothing\footnote{Under the alias of the
  ``Hodrick-Prescott filter'' \cite{Paige-Trindade-on-HP-filter}.}
\cite{DeJong-Dave-structural-macro}.

The fundamental problem is that in many complex, evolving systems, the
low-dimensional variables we happen to measure may develop in basically
unpredictable ways.  Old patterns may become completely irrelevant, even
actively misleading.  We could try to identify change-points and start modeling
afresh at each break, but there is often little {\em a priori} reason to think
that non-stationarities {\em will} take the form of abrupt breaks, as opposed
to more gradual transitions, to say nothing of all of the difficulties which
plague change-point detection.

While a non-stationary process {\em could} evolve in a totally capricious
fashion, more often there are at least local periods where the predictive
relationship between history and future does not change too rapidly.  For
stationary processes, this relationship is fixed and can be learned
nonparametrically
\cite{Ornstein-Weiss-how-sampling-reveals-a-process,Algoet-universal-schemes},
leading to forecasts with low risk, i.e., low expected loss on new data.  In
contrast, we follow the individual-sequence forecasting literature
\cite{prediction-learning-and-games} in wanting to have low \textbf{regret}
relative to a given collection of models --- no matter what sample path the
process realizes, we want to have done nearly as well the model, or sequence of
models, which in hindsight proves to have forecast best.  It is hard to see how
we could go beyond bounds on retrospective regret to bounds on prospective risk
in the face of arbitrary, unknown non-stationarities; to do so would be
tantamount to solving the problem of induction.

We start from algorithms for ``prediction with expert advice,'' which
adaptively combine the forecasts of an ensemble of models or ``experts'' so as
to guarantee low regret.  We focus on versions of the ``exponentially-weighted
average forecaster'' \cite{Littlestone-Warmuth-WM,Vovk-aggregating} (or
``multiplicative weight training''
\cite{Arora-Hazan-Kale-multiplicative-weights}), which forecasts a weighted
combination of the predictions of the experts in its ensemble, with weights
being multiplied up or down as experts do better or worse than the ensemble
average.  Using $\ensemblesize$ experts over $n$ rounds, this guarantees a
regret, with respect to the retrospectively-best single expert, of no more than
$O(\sqrt{n \ln{\ensemblesize}})$.

If instead of combining individual experts, we combine {\em sequences} drawn
from an ensemble of base experts, we can ``track the best expert''.
Specifically, the regret compared to a sequence where the base expert is
switched at most $m$ times follows the same form, but with the number of such
sequences in place of the number of base experts $\ensemblesize$; some
combinatorics \cite{prediction-learning-and-games} gives a bound of
$O(\sqrt{n\left((m+1)\ln{\ensemblesize} + (n-1)H(\frac{m}{n-1})\right)})$, with
$H(p) = -p\ln{p}-(1-p)\ln{(1-p)}$ being the binary entropy function, appearing
here via Stirling's formula.  The ``fixed shares'' algorithm introduced by
\cite{Herbster-Warmuth-tracking-the-best} implements this with only
$\ensemblesize$ weights, not a combinatorially-large number.

These algorithms are not quite suitable for the problems we have in mind,
however, because they presume that all experts in the ensemble are present at
the start.  Low regret relative to such an ensemble is not very comforting:
none of the experts might be much good, because one is faced with conditions
very different from any anticipated when the experts were set up.  One could
allow each expert to adapt --- rather than being a fixed forecasting rule,
regard each expert as estimating some statistical model from (some part) of the
sequence, and then forecasting on that basis.  This actually requires {\em no}
change to results for, for example, the fixed-share forecaster (see below),
because the conditions of the theorems put no limit on how the experts'
forecasts depend on the past, just that they do (measurably).

Our proposal therefore is to {\em grow} the ensemble, adding a new expert every
$\tau$ time-steps.  To cope with non-stationarity, which would mean that old
data becomes irrelevant, the expert added at time $k\tau$ is fitted to the data
from $(k-1)\tau+1$ onwards\footnote{The very first expert is initialized with
  some default parameter setting.}, and thereafter is free to keep on updating
its parameter estimates and, of course, its predictions, using new data.  As
the ensemble grows, the oldest model is always fitted to the complete time
series, followed by successively younger models which omit more and more of the
oldest data, until the very youngest model only fits to the last $\tau$ steps
or less.  There is thus always an expert which is fitted to the whole data
stream, and at the other extreme an expert fitted only to the most recent data.
If we can prove a regret bound for this growing ensemble, we will have
something which performs (nearly) as well as a rule which uses the whole of the
data, presumably optimal in the stationary case, as well as performing (nearly)
as well as an expert using only the last $\tau$ observations, as would be
suitable in case of a profound change-point or structural break.

We show how to modify the fixed shares algorithm to efficiently work with such
an ensemble, while still providing an $o(n)$ bound on tracking regret.  \S
\ref{sec:setting} fixes the setting and notation.  \S
\ref{sec:growing-ensemble} introduces the exponentially-weighted forecaster
over expert sequences drawn from a growing ensemble, and our modification of
the fixed-shares forecaster.  The major results, the equivalence of the two
forecasters and the regret bound, follow in \S\S
\ref{sec:equivalence}--\ref{sec:regret-bounds}.  \S \ref{sec:example} presents
an empirical example from macroeconomics.  \S \ref{sec:discussion} contrasts our
approach with previous work, and discusses its methodological significance.

\section{Setting and Notation}
\label{sec:setting}

We follow the usual setting of individual-sequence forecasting
\cite{prediction-learning-and-games}.  At each discrete time $t \in 1, 2,
\ldots n$, Nature produces an observation $y_t \in \mathcal{Y}$.  Nature may be
deterministic, stochastic, or even a clever and deceitful Adversary.  Our
forecaster has access to a set of experts (for us, a set depending on $t$),
with the $i^{\mathrm{th}}$ expert predicting $f_{i,t} \in \mathcal{D}$.  (The
``prediction'' could be an action, but for concreteness we will only talk about
predictions.)  The forecaster also has available the data $y_1, y_2, \ldots
y_{t-1}$, and combines this, along with the advice of the experts, to give a
prediction $\hat{p}_t \in \mathcal{D}$.  After the forecaster makes its
prediction, it learns $y_t$, leading to losses $\ell(f_{i,t},y_t)$ for the
experts and $\ell(\hat{p}_t,y_t)$ for the forecaster.

The aim of the forecaster is to have predicted almost as well as the best
expert, or even the best sequence of experts, no matter what Nature does.  The
{\bf tracking regret} of the forecaster with respect to a sequence of experts
$i_1, i_2, \ldots i_n$ is the difference in their cumulative losses:
\begin{equation}
R(i_1, i_2, \ldots i_n) = \sum_{t=1}^{n}{\ell(\hat{p}_{t}, y_{t})} - \sum_{t=1}^{n}{\ell(f_{i_t,t},y_{t})}
\end{equation}
Good forecasting strategies have regrets which can be bounded uniformly over
both expert sequences and observation sequences $y_1, \ldots y_n$.  Ideally the
bound would be $o(n)$, so that the regret per unit time goes to zero; in that
case the forecaster is ``Hannan-consistent.''

Some convenient abbreviations: $y_s^t$ is the sub-sequence of observations
$y_s, y_{s+1}, \ldots y_{t-1}, y_t$, and likewise for other sequence variables.
Further abbreviate $\ell(f_{i,t},y_t)$ by $\ell(i_{t},y_t)$, and
$\sum_{r=s}^{t}{\ell(i_r, y_r)}$ by $\ell(i_s^t,Y_s^t)$.  Regret is then
\begin{equation}
R(i_1^n) = \ell(\hat{p}_1^n,y_1^n) - \ell(i_1^n,y_1^n)
\end{equation}

\subsection{The basic forecasters}

\paragraph{The exponentially weighted average forecaster}
\cite{Littlestone-Warmuth-WM,Vovk-aggregating} Given an ensemble of
$\ensemblesize$ experts, initial (positive) weights $w_{i,0}$, and a learning
rate $\eta > 0$, this forecaster predicts by a weighted average\footnote{If
  convex combinations over $\mathcal{D}$ do not make sense, we use a randomized
  forecaster, complicating the notation a little and requiring us to bound
  expected regret rather than actual regret.  (By Markov's inequality, low
  expected regret implies low realized regret with high probability.)},
\begin{equation}
\hat{p}_t = \frac{\sum_{i=1}^{\ensemblesize}{w_{i,t-1} f_i,t}}{\sum_{j=1}^{\ensemblesize}{w_{j,t-1}}}
\end{equation}
and updates the weights by
\begin{equation}
w_{i,t} = w_{i,t-1}e^{-\eta\ell(f_{i,t},y_t)}
\end{equation}
This can be seen as a version of reinforcement learning, or as Bayes's rule (if
$\ell$ is negative log-likelihood), or as the evolutionary replicator dynamic,
with time-dependent fitness function $e^{-\eta\ell(f_{i,t},y_t)}$ [removed for
anonymous submission].

As mentioned above, the regret of the EWAF is $O(\sqrt{n \ln{\ensemblesize}})$
\cite{prediction-learning-and-games}.  If each member of the EWAF's ensemble is
actually a sequence over some class of base experts, we get a forecaster which
can keep low regret even if the best expert to use changes; the cost, however,
is keeping around a combinatorially-large number of weights.  The fixed shares
forecaster, described next, achieves the same results with only one weight for
each base expert, by modifying the manner in which weights are updated.

\paragraph{The fixed shares forecaster}
\cite{Herbster-Warmuth-tracking-the-best} We have $\ensemblesize$ experts, each
with a time-varying weight, and the forecast is, as before, a convex
combination:
\begin{equation}
\hat{p}_{t} = \frac{\sum_{i=1}^{\ensemblesize}{w_{i,t-1} f_{i,t}}}{\sum_{j=1}^{\ensemblesize}{w_{j,t-1}}}
\end{equation}
Initially, all weights are equal, $w_{i,0} = 1/\ensemblesize$.  The update
equations are
\begin{equation}
w_{i,t} = (1-\alpha)v_{i,t} + \alpha \frac{\sum_{i=1}^{\ensemblesize}v_{i,t}}{\ensemblesize}
\end{equation}
where 
\begin{equation}
v_{i,t} = w_{i,t-1}e^{-\eta\ell(i,y_{t})}
\end{equation}
and $\alpha \in [0,1]$ is another control setting.  In words, weights update
almost exactly as in the EWAF, except that weight is shared so that no expert
ever falls below a fraction $\alpha$ of the total weight.  As shown in
\cite{Herbster-Warmuth-tracking-the-best}, this matches the behavior of
exponential weighting over expert sequences, provided the initial weights of
sequences are {\em not} all equal; the weights are in fact chosen to depend on
the number of times the sequence changes expert, peaking when the number of
switches is about $\alpha n$.

\section{Growing Ensemble Forecasters}
\label{sec:growing-ensemble}

\subsection{The Growing Ensemble} We start with a single expert.  We divide the
time series into ``epochs,'' each of length $\tau$, and add a new expert at the
beginning of each epoch.\footnote{Only trivial changes are required to begin
  with $\ensemblesize_0$ experts and add $c$ new experts every epoch.}  When
added, the new expert is trained only on the data in the previous epoch.  The
number of experts at time $t$ is $\ensemblesize_t = 1 + \lfloor t/\tau
\rfloor$.  By time $n$, when the ensemble has $\ensemblesize_n = 1+ \lfloor
\frac{n}{\tau}\rfloor$ experts, one is trained over all data from time 1 to
$n$, one on data from $1+\tau$ to $n$, one on $1+2\tau$ to $n$, and so on, down to one
trained on the last $n \mod \tau$ observations.  The hope is that this will let
us cope with abrupt structural breaks (within at most $\tau$ time-steps),
gradual drift, and, of course, actual stationarity.

\subsection{Exponentially-Weighted Averaging over the Growing Ensemble}
To obtain a low tracking regret, we wish to run EWAF over sequences of experts
from the growing ensemble, limiting it, of course, to only using experts which
are currently available.  During the first $\tau$ time steps, there is only one
expert, but either of two experts can be used at any time from $t=1+\tau$ to
$t=2\tau$, any of three experts from $t=1+2\tau$ to $3\tau$, etc.  Even
limiting ourselves to sequences which switch experts no more than $m$ times
still leaves a combinatorially-large number of base-expert sequences, though
smaller than what would be the case if all $\ensemblesize_n$ final experts were
available from the beginning.

We will write the weight of the expert sequence $i_1^n$ at time $t$ as
$\ewaweight_t(i_1^n)$.  It is of course
\begin{equation}
\ewaweight_t(i_{1}^n) = \ewaweight_{t-1}(i_1^n)e^{-\eta\ell\left(i_t,y_t\right)} = \ewaweight_0(i_{1}^n)e^{-\eta\ell\left(i_{1}^t, y_{1}^t\right)}
\label{eqn:ewa-updating}
\end{equation}
We may regard $\ewaweight_t$ as a measure on the space of expert sequences of
length $n$, which defines measures on sub-sequences by summation; by a slight
abuse of notation we will also write them as $\ewaweight_t$, so
$\ewaweight_t(i_s^t) = \sum_{i_1^{s-1},i_{t+1}^n}{\ewaweight_t(i_1^{s-1},
  i_s^t, _{t+1}^n)}$.

We propose the following scheme of initial weights $\ewaweight_0(i_1^n)$.  Its
main virtue is that it can be emulated by a direct modification of the
fixed-shares forecaster, described immediately below.  We prove the emulation
result as Theorem \ref{thm:fixed-shares-emulates-compound-experts}.
\begin{equation}
 \frac{\ewaweight_0(i_{1}^{t + 1})}{\ewaweight_0(i_{1}^{t})} 
 =  \left\{ 
\begin{array}{ll}
  & 0 ~\mathrm{if} ~ i_{t+1} > \ensemblesize_{t+1}\\
&  \beta_t ~\mathrm{if} ~ t\bmod\tau = 0 ~\mathrm{and}~ i_{t+1} = \ensemblesize_{t+1}\\
&  \frac{\alpha}{\ensemblesize_{t+1}} + (1 - \alpha)\mathbf{1}_{\{ i_{t + 1} = i_{t}\}} ~\mathrm{otherwise}
 \end{array} 
 \right.
\end{equation}
That is, $\beta_t$ controls the weight assigned to an expert when it enters the
ensemble (and has no track record of losses).  The choice of $\beta$ is an
important issue, to which we return in the conclusion.  For the rest of this
paper, however, we set $\beta = \frac{\alpha}{\ensemblesize_{t+1}}$, so that
\begin{equation}
\frac{\ewaweight_0(i_{1}^{t + 1})}{\ewaweight_0(i_{1}^{t})} 
= \frac{\alpha}{\ensemblesize_{t+1}} + (1 - \alpha)\mathbf{1}_{\{ i_{t + 1} = i_{t}\}}.
\end{equation}
We abbreviate $\mathbf{1}_{\{ i_{t + 1} = i_{t}\}}$ by $\chi_t$, suppressing
explicit dependence on the sequence of actions.  Setting the base condition
$\ewaweight_0(1) = 1$ (because every sequence must begin with the single expert
available at the start), this recursively defines the initial weights for all
sequences of experts:
\begin{eqnarray}
\label{eqn:initial-initial-weight}
  \ewaweight_0(1) & = & 1\\
  \label{eqn:initial-weight-recursion}
  \ewaweight_0(i_1^{t+1}) & = & \ewaweight_0(i_1^t)\left(\frac{\alpha}{\ensemblesize_{t+1}} + (1-\alpha)\chi_t\right)
\end{eqnarray}
with the restriction $i_t \leq \ensemblesize_t$ understood.

\subsection{Growing-ensemble fixed shares forecaster}

The number of sequences of length $n$ from the growing ensemble is too large to
keep track of weights for each one, so, following the lead of
\cite{Herbster-Warmuth-tracking-the-best}, we introduce a fixed-shares
procedure which will turn out to match the weights induced by Eqs.\
\ref{eqn:ewa-updating} and \ref{eqn:initial-weight-recursion}.

At time $t$, each of the $\ensemblesize_t$ experts has a weight
$\fsweight_{i,t}$.  Initially, $\fsweight_{1,0} = 1$.  Thereafter, weights
update following the static ensemble procedure almost exactly.  For $ 1 \leq i \leq \ensemblesize_t$,
\begin{eqnarray}
\label{eqn:growing-ensemble-fixed-shares-update-1}
v_{i,t} & = & w_{i,t-1}e^{-\eta\ell(i,y_{t})}\\
\label{eqn:growing-ensemble-fixed-shares-update-2}
\fsweight_{i,t} & = & (1-\alpha)v_{i,t} + \frac{\alpha}{\ensemblesize_t} \sum_{i=1}^{\ensemblesize_t}{v_{i,t}}
\end{eqnarray}
and $w_{i,t} = 0$ for $i > \ensemblesize_t$.  Prediction, as always, is
a convex combination, $\hat{p}_t = \sum_{i=1}^{\ensemblesize_t}{\fsweight_{i,t-1} f_{i,t}}/\sum_{j=1}^{\ensemblesize_t}{\fsweight_{j,t-1}}$.

\subsection{Equivalence of Fixed Shares and Exponentially-Weighted Sequences}
\label{sec:equivalence}

Following \cite{Herbster-Warmuth-tracking-the-best}, we show that the fixed
shares algorithm assigns the same weight to any given base expert, at any given
time, as it gets from the exponentially-weighted averaging forecaster applied
to base-expert sequences.  This implies that they have the same behavior,
and in particular the same regret bounds.  Our proof is based on
that from \cite[Theorem 5.1, p.\ 103]{prediction-learning-and-games}.

\begin{theorem}
  Let $\ewaweight_{j,t} = \sum_{i_1^n: i_{t+1} =j}{\ewaweight_t(i_1^n)}$.  If
  $\ewaweight_0$ is set by Eq.\ \ref{eqn:initial-weight-recursion}, and
  $\fsweight$ updates by Eqs.\ \ref{eqn:growing-ensemble-fixed-shares-update-1}--\ref{eqn:growing-ensemble-fixed-shares-update-2},
  then for all $j$ and $t$ and $y_1^n$, $\ewaweight_{j,t} = \fsweight_{j,t}$.
  \label{thm:fixed-shares-emulates-compound-experts}
\end{theorem}

{\em Comment:} Recall that the EWAF will use $\ewaweight_{t-1}$, not
$\ewaweight_t$, to make its forecast at time $t$.

\paragraph{Proof} By induction on $t$.  When $t=0$, by construction,
$\fsweight_{1,0} = 1$, and $\fsweight_{j,0} = 0$ for all $j > 1$.  But this is
true for $\ewaweight_{j,0}$ as well, by Eq.\ \ref{eqn:initial-initial-weight}.
For the inductive step from $t-1$ to $t$, assume $\fsweight_{j,s} =
\ewaweight_{j,s}$ for all $j$ and for all $s < t$.  Write $\ewaweight_{i,t}$ as
a ``sum over histories'', using Eq.\ \ref{eqn:ewa-updating}.
\begin{eqnarray*}
\ewaweight_{i,t} 
& = & \sum_{i_{1}^{t}, i_{t + 2}^{n}}{ \ewaweight_t(i_{1}^{t}, i, i_{t+2}^{n})}
 =   \sum_{i_{1}^{t}}{e^{- \eta \ell\left(i_{1}^t,Y_{1}^t\right)} \ewaweight_0(i_{1}^{t}, i)}
 =  \sum_{i_{1}^{t}}{e^{- \eta \ell\left(i_{1}^t,Y_{1}^t \right)} \ewaweight_0(i_{1}^{t}) \frac{\ewaweight_0(i_{1}^{t}, i)}{\ewaweight_0(i_{1}^{t})}}\\
& = &  \sum_{i_{1}^{t}} {e^{- \eta \ell\left(i_{1}^t,Y_{1}^t \right)} \ewaweight_0(i_{1}^{t})\left(\frac{\alpha}{\ensemblesize_{t}} + (1 - \alpha)\chi_t \right)}
\end{eqnarray*}
by Eq.\ \ref{eqn:initial-weight-recursion}.  Moving the losses through time
$t-1$ into the weights,
\begin{eqnarray*}
& = & \sum_{i_{1}^{t}}{\ewaweight_{t - 1}(i_{1}^{t})e^{- \eta\ell\left(i_{t}, y_{t}\right)}\left(\frac{\alpha}{\ensemblesize_{t}} + (1 - \alpha)\chi_t \right)}
 =  \sum_{i_{t}}{\ewaweight_{i_{t}, t-1} e^{-\eta\ell\left(i_{t}, y_{t}\right)}\left(\frac{\alpha}{\ensemblesize_{t}} + (1 - \alpha)\chi_t \right)}\\
&= & \sum_{i_{t}}{\fsweight_{i_{t}, t-1} e^{- \eta\ell\left(i_{t}, y_{t}\right)}\left(\frac{\alpha}{\ensemblesize_{t}} + (1 - \alpha)\chi_t \right)}
\end{eqnarray*}
where we replace $\ewaweight_{i_{t}, t-1}$ with $\fsweight_{i_{t}, t-1}$ by the
inductive hypothesis.
\begin{equation}
 = \sum_{i_{t}}{v_{i,t}\left(\frac{\alpha}{\ensemblesize_{t}} + (1 - \alpha)\chi_t \right)} =
\fsweight_{i,t}
\end{equation}
by Eqs.\ \ref{eqn:growing-ensemble-fixed-shares-update-1}--\ref{eqn:growing-ensemble-fixed-shares-update-2}. $\Box$

\subsection{Regret bounds for the modified forecasters}
\label{sec:regret-bounds}

The {\bf size} $\sigma(i_1^n)$ of a sequence of experts is the number of times
the expert used changes, $\equiv \sum_{t=1}^{n-1}{\chi_t}$.  We require this to
be $\leq m$.  Let $m_k$ be the number of switches within the $k^{\mathrm{th}}$
epoch, i.e., $\sigma(i_1^{k\tau}) - \sigma(i_1^{(k-1)\tau})$, so
$\sum_{k=1}^{\ensemblesize_n}{m_{k}} = m$.

\begin{theorem}
  For all $n \geq 1$ and $y_1^n$, the tracking regret of the growing ensemble
  fixed shares forecaster is at most
  \begin{equation}
    R(i_{1}^n)  
    \leq
    \frac{m}{\eta}\ln{\ensemblesize_n} - \frac{1}{\eta}\ln{\alpha^{m}(1-\alpha)^{n-m}} + \frac{\eta}{8}n
  \end{equation}
  for all expert sequences $i_{1}^{n}$ where $m = \sigma(i_{1}^{n})$.
\label{thm:main-regret-bound}
\end{theorem}

\textsc{Proof} (after \cite{prediction-learning-and-games}, Theorem 5.2): The
key observation, proved as e.g. Lemma 5.1 in
\cite{prediction-learning-and-games}, is a general bound for
exponentially-weighted forecasters with unequal initial weights, which relates
their loss to the sum of the weights:
\begin{equation}
  \ell(\hat{p}_{1}^{t}, y_{1}^{t}) 
  \leq  -\frac{1}{\eta}\ln{\sum_{i_1^n}{\ewaweight_{n}(i_1^n)}} + \frac{\eta}{8}n
\end{equation}
Since weights are non-negative and $\ln$ is an increasing function, this implies
\begin{equation}
\ell(\hat{p}_{1}^{t}, y_{1}^{t}) \leq -\frac{1}{\eta}\ln{\ewaweight_n(i_{1}^{n})} + \frac{\eta}{8}n
\label{eqn:mixture-loss-in-terms-of-final-weights}
\end{equation}
for any action sequence $i_1^n$.  By the construction of the exponentially
weighted forecaster,
\begin{equation}
\ln{\ewaweight_n(i_{1}^{n})} = \ln{\ewaweight_0(i_{1}^{n})} - \eta\ell(i_{1}^n, y_{1}^n)
\label{eqn:final-weights}
\end{equation}
Assuming $\sigma(i_1^n) \leq m$, the initial log weight is bounded by
construction:
\begin{equation}
\ewaweight_0(i_{1}^{n}) 
=   \prod_{k = 1}^{\ensemblesize_n}{{\left(\frac{\alpha}{k}\right)}^{m_k}{\left(\frac{\alpha}{k} + 1 - \alpha\right)}^{\tau - m_{k}}}
 \geq  \prod_{k = 1}^{\ensemblesize_n}{{\left(\frac{\alpha}{k}\right)}^{m_{k}}{(1 - \alpha)}^{\tau - m_{k}}}
 \geq  {\left(\frac{\alpha}{\ensemblesize_n}\right)}^{m}(1 - \alpha)^{n-m}
\label{eqn:initial-weights-bound}
\end{equation}
Substituting Eq.\ \ref{eqn:initial-weights-bound} into Eq.\
\ref{eqn:final-weights}, and the latter into Eq.\
\ref{eqn:mixture-loss-in-terms-of-final-weights}, we get that
\begin{equation}
  \ell(\hat{p}_{1}^{t}, y_{1}^{t}) \leq \ell(i_{1}^n, y_{1}^n) + \frac{m}{\eta}\ln{\ensemblesize_n} - \frac{1}{\eta}\ln{\alpha^{m}(1-\alpha)^{n-m}} + \frac{\eta}{8}n
\end{equation}
and the theorem follows. $\Box$

The familiar regret bound for the exponentially-weighted forecaster with equal
initial weights is that $R$ is at most $O(\sqrt{n\ln{N}})$, with $N$ being the
size of the ensemble.  Since the number of allowable expert sequences of length
$n$ with $m$ switches is at most $(\ensemblesize_n)^m {n-1 \choose m}$, we
would be doing well to achieve a regret bound of $O(\sqrt{n ( \log{{n-1 \choose
      m}} + m \ln{\ensemblesize_n})})$.  This can in fact be done by tuning
$\alpha$ and $\eta$.


\begin{corollary}
Fix $n$ and $m$, and run the modified fixed share forecaster with $\widehat{\alpha} = \frac{m}{n-1}$, and
\begin{equation}\textstyle
\widehat{\eta} = \sqrt{\frac{8}{n}\left((n-1)H(\widehat{\alpha}) - \ln(1-\widehat{\alpha}) + m\ln \ensemblesize_n \right)},
\end{equation}
then 
\begin{equation}
  R(i_{1}^{n}) \le \sqrt{\frac{n}{2}\left( (n-1)H\left(\widehat{\alpha}\right) - \ln{\left(1-\widehat{\alpha}\right)} +  m\ln{\ensemblesize_n} \right)}
\end{equation}
for any action sequence $i_{1}^{n}$ making at most $m$ switches.
\label{cor:optimal-control-settings}
\end{corollary}

\paragraph{Proof} (After \cite[Cor.\ 5.1, p.\
105]{prediction-learning-and-games}): Let $\alpha = \frac{m}{n-1}$. Then:
\begin{eqnarray*}
\ln{\left(\frac{1}{\widehat{\alpha}^{m}(1 - \widehat{\alpha})^{n-m}} \right)}
& = & - m \ln{\widehat{\alpha}} - (n - m)\ln{(1 - \widehat{\alpha})}\\
& = & -m\ln{\widehat{\alpha}} - (n-m-1)\ln{(1-\widehat{\alpha})} -\ln{(1-\widehat{\alpha})}\\
 &= & (n - 1)H(\widehat{\alpha}) - \ln{\left(1-\widehat{\alpha}\right)}.
\end{eqnarray*}

Substituting $\eta$ into the regret bound, and using this equality, we are done.
$\Box$

{\em Remark 1:} Notice that for fixed $m$, $\widehat{\alpha} \rightarrow 0$ as
$n\rightarrow\infty$, so that $H(\widehat{\alpha}) \rightarrow 0$ and the
over-all bound is $o(n)$.

{\em Remark 2:} The learning rate and minimum share could probably be tuned
better by more careful counting of the number of size-$m$ sequences from the
growing ensemble, but since this will only improve the comparatively-small
$m\ln{\ensemblesize_n}$ term, we omit the combinatorics here.

\section{Example: GDP Forecasting}
\label{sec:example}

We illustrate our approach by predicting a non-stationary time series of great
practical importance, the gross domestic product (GDP) of the United States,
recorded quarterly from the second quarter of 1947 to the first quarter of 2010
(from the FRED data service of the Federal Reserve Bank of St.\ Louis).  After
following the common practice of converting this to quarterly growth rates,
this gives $n=252$ observations.  Somewhat arbitrarily, we made all of our
models linear autoregressions of order 12 (i.e., AR(12) models), set the epoch
length $\tau$ to 16 quarters or 4 years, and allowed $m=15$ switches of expert,
with $\alpha$ and $\eta$ then following by Corollary
\ref{cor:optimal-control-settings}.

\begin{figure}
\begin{center}
$a$ \includegraphics[width=0.45\textwidth]{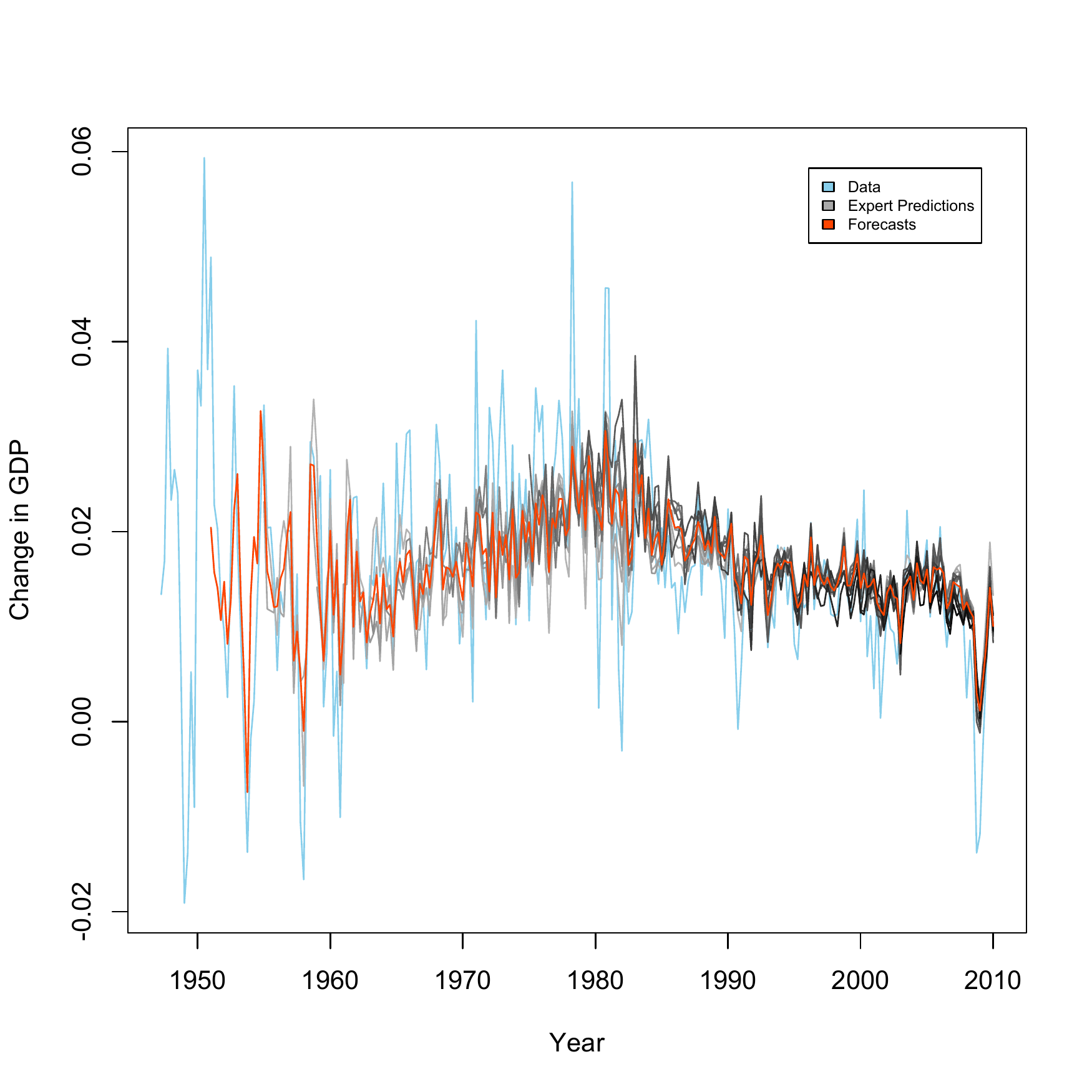}
$b$ \includegraphics[width=0.45\textwidth]{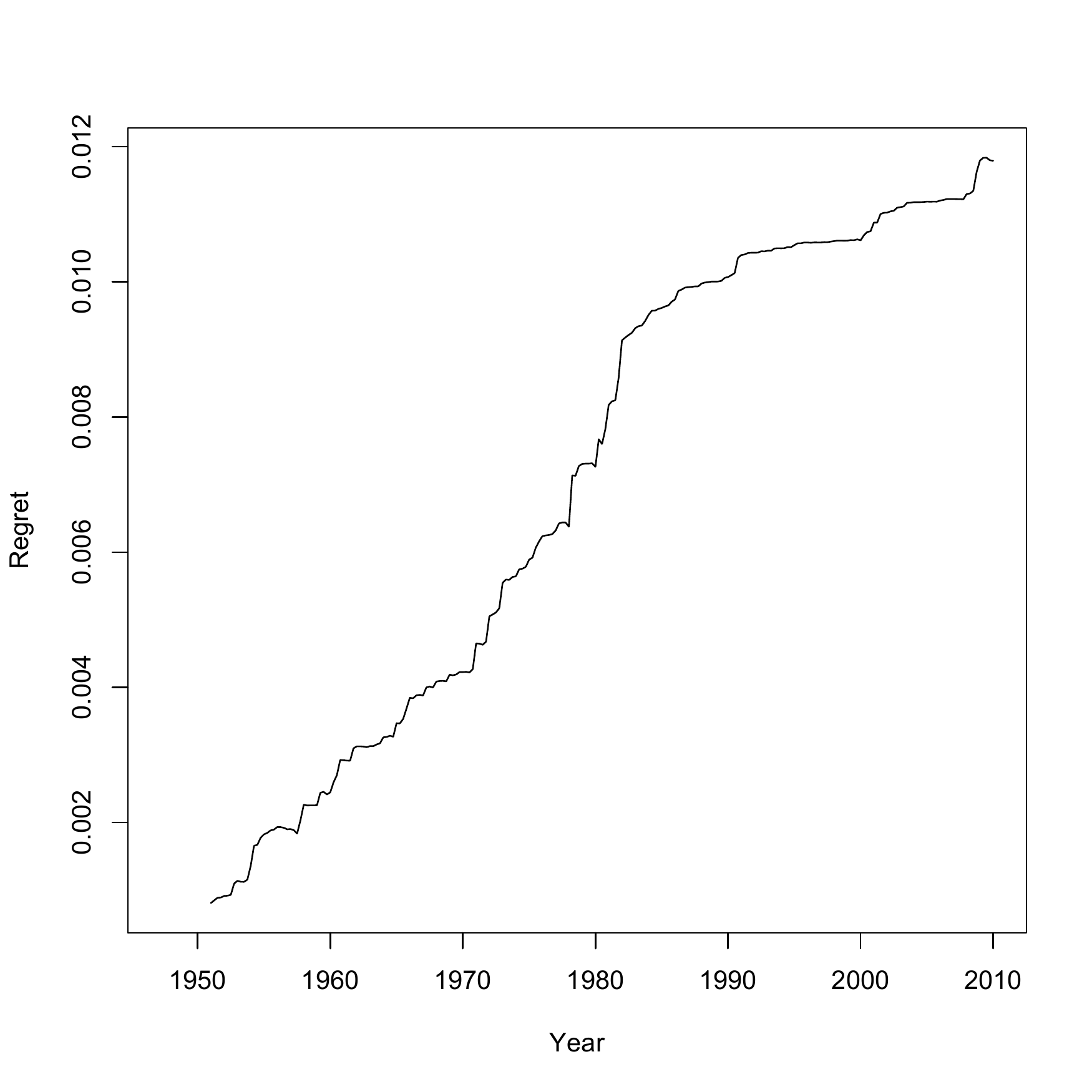}
\end{center}
\caption{$a$: Quarterly growth rate of US GDP, 1948--2010, with predictions of
  the growing ensemble of AR(12) models and the weighted ensemble forecast.
  $b$ Accumulated regret of the ensemble, compared to combining global spline
  smoothing with an ARMA(8,7) model.}
\label{fig:gdp}
\end{figure}

Figure \ref{fig:gdp}$a$ shows the evolution of GDP growth (clearly
non-stationary), as well as the evolution of the ensemble and its
weighted-average prediction, which does quite well despite the fact that AR
models are both the simplest possible predictors here, and are all assured
mis-specified\footnote{They are confidentaly rejected by Box-Ljung tests
  \cite{Shumway-Stoffer}.}.  State-of-the-art economic forecasts rely on
complicated multivariate state-space models called DSGEs
\cite{DeJong-Dave-structural-macro}, after de-trending with a smoothing spline.
For GDP, however, the predictions of DSGEs are close to those of a simple
autoregressive moving average (ARMA) model, so we fit one to spline residuals;
AIC order selection \cite{Shumway-Stoffer} gave us an ARMA(8,7).  Figure
\ref{fig:gdp}$b$ shows the accumulated loss of the ensemble compared to this
model; it is both small and growing sub-linearly, despite the fact that the
ARMA model has much more memory than the ARs (because of the moving-average
component), and it takes advantage of the flexibility of non-parametric (and
indeed non-causal) smoothing in the spline.  Calculating regret against the
best sequence of models from the ensemble, allowing $m=n$, produced a similar
profile over time (not shown), but even smaller comparative losses.

\section{Discussion}
\label{sec:discussion}

\subsection{Related Work}

The closest approach to our method is that of Hazan and Seshadhri \cite{Hazan-Seshadhri}, who also
work within the family of variants on multiplicative weight training.  They
introduce a new expert at each time step, whose initial weight is a fixed
function of time, and do not otherwise implement a ``fixed share'' of weights,
i.e., a minimum weight for each expert.  Maintaining such a fixed share is
extremely useful when a pre-existing model {\em becomes} one of the best,
drastically cutting the time needed for it to dominate the ensemble.
\cite{Hazan-Seshadhri} also does not use tracking regret, but rather the
maximum regret against any single expert attained over any contiguous time
interval.  This time-uniform regret is attractive, and they prove bounds on it,
but only by assuming that each individual expert itself has a low, time-uniform
regret (in the ordinary sense); some of their results even require low losses,
not just low regrets.  Our approach, by contrast, is able to accommodate the
much more realistic situation where each individual expert may indeed have high
loss, or even high regret, because the process is hard to predict and no one
model is uniformly applicable.

Turning to more conventional approaches, econometrics has a large literature on
detecting non-stationarity (of the basically-harmless ``integrated'' type
characteristic of random walks), and finding ``structural breaks'' (change
points), after which models must be re-estimated or re-specified
\cite{Clements-Hendry-forecasting-nonstationary}.  Economists do not seem to
have considered an ensemble method like ours, perhaps due to their laudable (if
unfulfilled) ambition to capture the exact data-generating process in a single
parsimonious model.  Similarly, most work on data-set shift and concept drift
in machine learning \cite{dataset-shift-in-ML} deals with how a single model
should be learned (or modified) so as to be robust to various changes in the
joint distribution of inputs and outputs.  Unlike all these approaches, we do
not have to assume that {\em any} of our models are well-specified, nor assume
anything about the nature of the data-generating process or how it changes over
time.

There are some ensemble methods which are reminiscent of aspects of our
proposal, such as Kolter and Maloof's ``additive expert ensemble'' algorithm
\texttt{AddExp} \cite{Kolter-Maloof-AddExp}, the incremental-learning SEA
algorithm \cite{Street-Yim-SEA}, and adaptive time windows algorithms
(e.g. \cite{Scholz-Klinkenberg-drifting-concepts}).  None of these allow the
full combination of a growing ensemble with temporally-specialized experts and
adaptive weights.  Consequently, while some of them can handle mild
non-stationarities if the base models are close to well-specified, none of them
are able to make strong individual-sequence prediction guarantees like those of
Theorem \ref{thm:main-regret-bound}.

\subsection{Conclusion}

We have introduced the growing-ensemble method, and shown that it leads to a
modification of the conventional fixed-shares forecaster which is still
Hannan-consistent, with $o(n)$ regret over $n$ time-steps compared to the
retrospectively-best sequence of experts.  This bound takes into account the
fact that the ensemble grows continually and that individual experts can be
arbitrarily bad, while the time series can have arbitrary non-stationarities.
There are several interesting technical directions in which to take this (Can
the counting of experts be replaced with variation of losses across the
ensemble, as in \cite{Hazan-Kale-variation-in-costs}?  Would it help to vary
the weight with which new experts get introduced? Is there an optimal epoch
length $\tau$?), the real importance of this work is methodological.

Complex systems tend to produce time series which are not just non-stationary
but genuinely evolutionary --- even if there is, in some sense, a fixed
high-dimensional generative model, the dynamics of the low-dimensional
variables we deal with changes in character over time.  Tractable prediction
models for such time series are at best local and transient approximations, no
single one of which will work well for long.  It is implausible even to come up
with a fixed collection of models before we see how the system actually
develops.  Our growing ensemble method accommodates arbitrary dynamics, without
assuming well-specified models, trends that can be extrapolated, stationary
behavior punctuated by well-defined structural breaks, or other such props
supporting previous work.  Giving up the desire for the One True Model, of
minimal risk, in favor of a growing ensemble of imperfect models, means we
adapt automatically to arbitrary, historically evolving non-stationarities ---
including stationarity.

\bibliography{locusts}

\begin{thebibliography}{10}

\bibitem{Algoet-universal-schemes}
Paul Algoet.
\newblock Universal schemes for prediction, gambling and portfolio selection.
\newblock {\em Annals of Probability}, 20:901--941, 1992.

\bibitem{Arora-Hazan-Kale-multiplicative-weights}
Sanjeev Arora, Elad Hazan, and Satyen Kale.
\newblock The multiplicative weights update method: a meta algorithm and
  applications, 2005.

\bibitem{Caires-Ferreira-prediction-of-cond-stat-seq}
S.~Caires and J.~A. Ferreira.
\newblock On the non-parametric prediction of conditionally stationary
  sequences.
\newblock {\em Statistical Inference for Stochastic Processes}, 8:151--184,
  2005.
\newblock Correction, vol. 9 (2006), pp. 109--110.

\bibitem{prediction-learning-and-games}
Nicol{\`o} Cesa-Bianchi and G{\'a}bor Lugosi.
\newblock {\em Prediction, Learning, and Games}.
\newblock Cambridge University Press, Cambridge, England, 2006.

\bibitem{Clements-Hendry-forecasting-nonstationary}
Michael~P. Clements and David~F. Hendry.
\newblock {\em Forecasting Non-stationary Economic Time Series}.
\newblock MIT Press, Cambridge, Massachusetts, 1999.

\bibitem{DeJong-Dave-structural-macro}
David~N. DeJong and Chetan Dave.
\newblock {\em Structural Macroeconometrics}.
\newblock Princeton University PRess, Princeton, New Jersey, 2007.

\bibitem{Hazan-Seshadhri}
Elad Hazan and C. Seshadhri.
\newblock Efficient learning in algorithms for changing environments.
\newblock In {\em Proceedings of the 26th Annual International Conference on Machine Learning [ICML 2009]}, pages 
393--400. ACM Press, 2009. 

\bibitem{Hazan-Kale-variation-in-costs}
Elad Hazan and Satyen Kale.
\newblock Extracting certainty from uncertainty: Regret bounded by variation in
  costs.
\newblock {\em Machine Learning}, 80:165--188, 2010.

\bibitem{Herbster-Warmuth-tracking-the-best}
Mark Herbster and Manfred Warmuth.
\newblock Tracking the best expert.
\newblock {\em Machine Learning}, 32:151--178, 1998.

\bibitem{Kolter-Maloof-DWM}
J.~Zico Kolter and Marcus~A. Maloof.
\newblock Dynamic weighted majority: An ensemble method for drifting concepts.
\newblock {\em Journal of Machine Learning Research}, 8:2755--2790, 2007.

\bibitem{Kolter-Maloof-AddExp}
Jeremy~Z. Kolter and Marcus~A. Maloof.
\newblock Using additive expert ensembles to cope with concept drift.
\newblock In {\em Proceedings of the 22nd International Conference on Machine
  Learning}, pages 449--456. ACM Press, 2005.

\bibitem{Littlestone-Warmuth-WM}
N.~Littlestone and Manfred Warmuth.
\newblock The weighted majority algorithm.
\newblock {\em Information and Computation}, 108, 1994.

\bibitem{Ornstein-Weiss-how-sampling-reveals-a-process}
Donald~S. Ornstein and Benjamin Weiss.
\newblock How sampling reveals a process.
\newblock {\em Annals of Probability}, 18:905--930, 1990.

\bibitem{Paige-Trindade-on-HP-filter}
Robert~L. Paige and A.~Alexandre Trindade.
\newblock The {Hodrick}-{Prescott} filter: A special case of penalized spline
  smoothing.
\newblock {\em Electronic Journal of Statistics}, 4:856--874, 2010.

\bibitem{dataset-shift-in-ML}
Joaquin Qui{\~n}onero-Candela, Masashi Sugiyama, Anton Schwaighofer, and
  Neil~D. Lawrence, editors.
\newblock {\em Dataset Shift in Machine Learning}.
\newblock MIT Press, Cambridge, Massachusetts, 2009.

\bibitem{Scholz-Klinkenberg-drifting-concepts}
Martin Scholz and Ralf Klinkenberg.
\newblock An ensemble classifier for drifting concepts.
\newblock In Porto ECML, editor, {\em Proceedings of the Second International
  Workshop on Knowledge Discovery and Data Mining (KDD)}, 2005.

\bibitem{Shumway-Stoffer}
Robert~H. Shumway and David~S. Stoffer.
\newblock {\em Time Series Analysis and Its Applications}.
\newblock Springer Texts in Statistics. Springer-Verlag, New York, 2000.

\bibitem{Street-Yim-SEA}
W.~N. Street and Y.~Kim.
\newblock A streaming ensemble algorithm {(SEA)} for large-scale
  classification.
\newblock In {\em Proceedings of the Second International Workshop on Knowledge
  Discovery and Data Mining {(KDD)}}, pages 377--382, New York, 2001. ACM
  Press.

\bibitem{Vovk-aggregating}
Volodimir Vovk.
\newblock Aggregating strategies.
\newblock In Mark Fulk and John Case, editors, {\em Proceedings of the Third
  Annual Workshop on Computational Learning Theory [COLT 1990]}, pages
  371--386, San Francisco, 1990. Morgan Kaufmann.

\end{thebibliography}
\bibliographystyle{plain}

\end{document}